\documentclass[11pt]{article}

\usepackage{dialogue}

\usepackage{ifxetex}
\ifxetex
    \usepackage{fontspec}
    \setromanfont{Times New Roman}
\else
  \usepackage{cmap}
  \usepackage[T2A,T1]{fontenc}
  \usepackage[utf8]{inputenc}
  \usepackage{times}
  \usepackage{latexsym}
  \usepackage{substitutefont}
  \substitutefont{T2A}{\familydefault}{cmr}
\fi

\usepackage[russian,british]{babel}

\usepackage[dvipsnames]{xcolor} %
\usepackage{booktabs}  %
\usepackage{multirow}  %
\usepackage{url}
\usepackage{pgf}
\usepackage[normalem]{ulem}  %

\usepackage{newtxtext,newtxmath}  %

\usepackage{covington} %

\usepackage{hyperref}

\dialogfinalcopy %

\title{Light Coreference Resolution for Russian \\with Hierarchical Discourse Features}

\author{Elena Chistova \and Ivan Smirnov \\
        FRC CSC RAS \\
        Moscow, Russia \\
        \tt{\{chistova, ivs\}@isa.ru} \\
        }

\date{}

\begin{document}

\maketitle
\begin{abstract}

Coreference resolution is the task of identifying and grouping mentions referring to the same real-world entity. Previous neural models have mainly focused on learning span representations and pairwise scores for coreference decisions. However, current methods do not explicitly capture the referential choice in the hierarchical discourse, an important factor in coreference resolution. In this study, we propose a new approach that incorporates rhetorical information into neural coreference resolution models. We collect rhetorical features from automated discourse parses and examine their impact. As a base model, we implement an end-to-end span-based coreference resolver using a partially fine-tuned multilingual entity-aware language model LUKE. We evaluate our method on the RuCoCo-23 Shared Task for coreference resolution in Russian. Our best model employing rhetorical distance between mentions has ranked 1st on the development set (74.6\% F1) and 2nd on the test set (73.3\% F1) of the Shared Task\footnote{The code and models are available at \url{https://github.com/tchewik/corefhd}}. We hope that our work will inspire further research on incorporating discourse information in neural coreference resolution models.
  
  
  \textbf{DOI:} 10.28995/2075-7182-2022-20-XX-XX
\end{abstract}

\selectlanguage{russian}
\begin{center}
  \russiantitle{Разрешение кореференции для русского языка \\с использованием признаков иерархического дискурса}

    \medskip \setlength\tabcolsep{0.5cm}
    \begin{tabular}{cc}

    \multicolumn{2}{c}{\textbf{Чистова Е. В.,  Смирнов И. В.}} \\
    \multicolumn{2}{c}{ФИЦ ИУ РАН}                 \\
    \multicolumn{2}{c}{Москва, Россия}             \\
    \multicolumn{2}{c}{\{chistova, ivs\}@isa.ru}  
    \end{tabular}
    \medskip
  
\end{center}

\begin{abstract}
  Разрешение кореференции -- это задача выявления и группировки упоминаний, относящихся к одному и тому же объекту реального мира. При решении задачи методами глубокого обучения в первую очередь обращают внимание на проблемы обучения векторных представлений сущностей и оценки вероятности наличия референциальной связи между ними. Однако существующие методы не позволяют в явном виде учитывать референциальный выбор в иерархическом дискурсе. В данной работе оценивается важность признаков, полученных на основе автоматического риторического анализа, применительно к нейросетевым моделям. В качестве базового метода реализована end-to-end архитектура с использованием мультиязычной языковой модели LUKE, учитывающей при кодировании текста границы сущностей. Лучшая модель, в которой используется признак риторического расстояния между сущностями, занимает первое место на валидационной (74.6\% F1) и второе место на тестовой (73.3\% F1) выборке соревнования RuCoCo-2023. 
  
  \textbf{Ключевые слова:} разрешение кореференции, теория риторических структур, референциальный выбор, риторическое расстояние, русский язык
\end{abstract}
\selectlanguage{british}

\section{Introduction}
\label{intro}

Coreference resolution is the task of identifying and grouping mentions referring to the same real-world entity. It is a challenging task in natural language processing, as it often requires both linguistic and common knowledge. In recent years, neural models have achieved remarkable success in coreference resolution. These models aim to identify mention spans and assign pairwise scores. However, they mostly rely on surface explicit features, such as the distance between entities in tokens, and overlook the hierarchical discourse structure. Contextual word embeddings, despite their morphosyntactic and semantic richness, also have limitations in capturing document discourse beyond local cues.

Our system for RuCoCo-2023, called CorefHD (\textbf{Coref}erence in \textbf{H}ierarchical \textbf{D}iscourse), enhances the classical neural architecture with automatically retrieved features that capture aspects of hierarchical discourse. It uses pretrained transformer-based contextualized word embeddings, along with dense embeddings of hierarchical discourse features: linear distance, rhetorical distance, and anaphor-to-LCA distance. To retrieve the discourse hierarchy of the text, we use an RST parser predicting constituency trees in accordance with the Rhetorical Structure Theory \cite{mann1988rhetorical}.

The main contributions of this paper are:

\begin{itemize}
\item We propose a new method that incorporates discourse information into neural coreference resolution models.
\item We test various discourse features that capture the distances between mentions on a large coreference resolution dataset in Russian.
\item We apply a number of memory reduction techniques and demonstrate that high-quality coreference resolution can be done with standard neural architecture even with limited computational resources.
\item We use a multilingual entity-aware LUKE \cite{yamada-etal-2020-luke} language model and show that it performs competitively with the monolingual language models for Russian in coreference resolution.
\item We join the RuCoCo-2023 Shared Task, and achieve 1st place on the development set and 2nd place on the test set of the contest with the model using the rhetorical distance feature.
\end{itemize}

The rest of this work is organized as follows: Section \ref{sec:related_work} reviews a concept of referential distance and current work on coreference resolution in hierarchical discourse. Section \ref{sec:approach} describes our method in detail. Section \ref{sec:experiments} presents our experimental setup. Section \ref{sec:results} analyzes our results. Section \ref{sec:conclusion} concludes the paper and discusses future work.

\section{Related Work}
\label{sec:related_work}

Linear referential distance measures how many clauses separate an anaphor from its antecedent \cite{givon1983topic}. However, not all phrases in discourse require the same level of attention. It is observed \cite{grosz-sidner-1986-attention} that the discourse structure of a text contains discourse units inside and outside the intention and attention. Using a corpus of 30 manually annotated texts, it is shown \cite{cristea-etal-1999-discourse} that a hierarchical model of discourse has greater potential for improving the coreference resolution performance than a linear model of discourse. The most popular hierarchical discourse framework as of today is Rhetorical Structure Theory \cite{mann1988rhetorical}. Within RST, one can consider in the referential distance the rhetorical structures, where attention focus is part of the definition \cite{moser-moore-1996-toward} of subordinating (mononuclear) RST relations. An approach to computing referential distance with respect to the rhetorical tree is suggested by Kibrik \cite{kibrik1999cognitive}: the rhetorical distance can be measured by counting the nodes in an RST tree that are visited while walking from the mention to its possible antecedent. A study on the RST Discourse Treebank\footnote{\url{https://catalog.ldc.upenn.edu/LDC2002T07}} shows that while rhetorical distance does not imply the one and only referential choice, it is still one of the principal factors for referential choice prediction \cite{kibrik2005corpus}. Another study \cite{fedorova-etal-2010-experimental} uses six RST-annotated text fragments in Russian to demonstrate that rhetorical distance has a significant impact on the referent activation in working memory.

Closest to our work are \cite{khosla-etal-2021-evaluating} implementing various features over an RST tree produced with a parser for English. However, their main concern is how general is the lowest common ancestor of two mentions in the rhetorical constituency tree. While this is somewhat related to the working memory load of keeping two mentions active, they do not directly consider a concept of referential distance and, most importantly, ignore nuclearity (i.e. attention), which is a crucial feature in rhetorical structures.

In this paper, we apply the RST parser for Russian to build hierarchical discourse trees. The distance features obtained from these trees we use in a neural coreference resolution model. As far as we know, we are the first to model referential distances in hierarchical discourse with neural models. We also examine the impact of the RST features in coreference resolution for Russian on a large annotated corpus.

\section{Approach}
\label{sec:approach}

End-to-end coreference resolution involves finding entities in plain text and collecting them into clusters so that each cluster corresponds to a single real-world object.  As a core method, we apply the classical \cite{lee-etal-2018-higher}'s approach to end-to-end coreference parsing with a span-ranking architecture, except for the higher-order inference which has been proven to be ineffective \cite{xu-choi-2020-revealing}. This approach to coreference resolution involves five main steps: 
\begin{enumerate}
    \item Collect the initial set of spans.
    \item Rank the collected spans with a linear transformation of span embeddings and keep the top-k resembling entities.
    \item Collect the coarse referent-to-antecedent probabilities for each possible pair of entities. This is calculated as a sum of corresponding span probabilities obtained in the previous step and a score obtained with a bilinear transformation of two \textit{mention encodings}. Keep the top-n pairs with the highest prediction.
    \item Compute the final coreference scores for each possible mention-antecedent pair that made it to this step. This is done with a feedforward layer processing \textit{mention pair encodings}. Assign to each mention the antecedent with the highest predicted probability.
    \item The predictions form connected chains of mentions that can be viewed as clusters.
\end{enumerate}

\vspace{3mm}
\noindent The following gives the details of how our system encodes entities and their pairs.

\paragraph{Mention Encoding} Each fine-grained token is encoded as an average of its subtoken representations obtained using a language model. The initial entity candidates are collected greedily, with the only parameter being the maximum length of the span. To adjust this parameter effectively, we use token representations instead of LM subtoken representations. Since language models work with a limited context, we collect each paragraph representation separately.

\paragraph{Mention Pair Encoding} To calculate the final predictions for each pair of found mentions, we use a feedforward layer that takes a mention pair embedding as input. This embedding consists of the concatenation of two individual mention encodings and the embedding of the token count between them. For the models employing discourse hierarchy features, we represent them similarly to token distances and concatenate them to the pair embeddings.

\subsection{Discourse Hierarchy Features}
\label{sec:hd_features}

Given two spans $i$ (a mention) and $j$ (its possible antecendent), we first find the elementary discourse units $u_i$ and $u_j$ covering the corresponding spans in a predicted RST tree. Then we compute the discourse-related features and concatenate them with mention pair encoding.

Two metrics are used to measure referential distance in discourse, as outlined in \cite{kibrik1999cognitive}:

\begin{itemize}

\item \textbf{Linear Distance ($D_{Lin}$)} in our model is a number of predicted elementary discourse units (EDUs) occurring between two spans. 

\item \textbf{Rhetorical Distance ($D_{Rh}$)} is a number of nuclear EDUs occurring between two spans in a hierarchical rhetorical tree. %

\end{itemize}

We also adopt a feature estimating the amount of generality required to have two mentions in the same discourse subtree \cite{khosla-etal-2021-evaluating}:

\begin{itemize}
\item \textbf{Referent's distance to the LCA ($D_{LCA}$)} Assuming mention $i$ always appearing to the right of any possible antecedent~$j$, and $LCA(u_j, u_i)$ being the lowest discourse unit covering both $u_i$ and $u_j$ in the constituency RST tree, $D_{LCA} = \textrm{dist}(u_i,~LCA(u_j, u_i))$.
\end{itemize}

\section{Experimental Setup}
\label{sec:experiments}

\subsection{Pretrained Language Model}
\label{ssec:pretrained_lm}
We employ the multilingual LUKE\footnote{\texttt{studio-ousia/mluke-large-lite}} \cite{ri-etal-2022-mluke}. It is a language model that has been trained with both masked language modeling (MLM) and masked entity prediction (MEP) tasks. The entity annotations in the training corpus are collected from hyperlinks in Wikipedia dumps. This multilingual model has previously demonstrated significant improvement in question answering and cloze prompt tasks for Russian compared to mBERT \cite{devlin-etal-2019-bert} and XLM-RoBERTa \cite{conneau2019cross}. We hypothesize that explicit coreference resolution can also benefit from LM-ingrained entity encoding.

\subsection{Factors Reducing Memory Consumption}
\label{ssec:factors_reducing_memory}

Neural coreference resolution is a memory-intensive task. The common approach to end-to-end coreference resolution \cite{lee-etal-2017-end,lee-etal-2018-higher} requires computation over each and every span in a document. A number of recent works suggest more optimal alternative methods, in which the object of processing is not a span but a token \cite{kirstain-etal-2021-coreference,thirukovalluru-etal-2021-scaling,dobrovolskii-2021-word}. Despite this, the relevant research adopting language model fine-tuning still requires 40 to 80 GB of video memory \cite{dobrovolskii-2021-word,maehlum-etal-2022-narc}. In our study, we investigate the extent to which the most classical span-based approach to coreference resolution can be scaled down.

Each our model is trained on a single 32GiB Tesla V100 GPU, with peak memory allocation of 98\%. To achieve this, we modified the standard model architecture and implementation:
\begin{itemize}
    \item The main factor that allows a coreference model to be trained on a large dataset with limited memory is {\it excluding full LM fine-tuning}. In our experiments, a language model is frozen except for the last $k$ layers. The value of $k$ is determined empirically by the amount of video memory available. In our setting, $k=8$ of 23 layers.
    \item After initial token encoding, the bidirectional LSTM is used to obtain {\it lower-dimensional token embeddings}. The span embedding is computed from the sequence of compressed token embeddings using self-attention. In our experiments, $\mathbf{e}_{LM} \in \mathbb{R}^{1024}$ and $\mathbf{e}_{LSTM} \in \mathbb{R}^{100}$.
    \item {\it Each paragraph of the text is encoded with a language model separately.} This allows long news articles to be encoded without trimming and high-dimensional partially-trainable LM embeddings to be compressed in place, thereby saving memory.
    
\end{itemize}

We also use standard techniques reducing memory requirements:
\begin{itemize}
    \item {\it Batch size} = 1. Gradient accumulation did not improve training results.
    \item All the calculations are performed with {\it mixed precision}.
\end{itemize}

\subsection{Instruments for Linguistic Analysis}
\label{ssec:instruments}

Tokenization and sentence splitting are performed with the Razdel\footnote{\url{https://github.com/natasha/razdel}} library. Named entities are recognized with the SpaCy\footnote{\url{https://spacy.io/}} \texttt{ru\_core\_news\_lg} model predicting BIO-tags from token embeddings. Discourse structures are produced with the IsaNLP RST\footnote{\url{https://github.com/tchewik/isanlp\_rst}} parser for Russian \cite{chistova2021rst}. The parser generates trees for each paragraph; we merged these trees with a right-branching multinuclear \textsc{Joint} relation to construct the full-text RST trees.

\subsection{Data}
\label{ssec:data}

\begin{table}
\small
\begin{tabular}{p{0.48\textwidth}p{0.48\textwidth}}
\toprule
\bf Original  & \bf Translation \\
\midrule

\foreignlanguage{russian}{Обитатели небоскребов Нью-Йорка спешат обзавестись [\textcolor{Emerald}{парашютами}]$_{SA}$

Обитатели небоскребов Нью-Йорка спешат обзавестись [\textcolor{Emerald}{парашютами}]$_{SA}$. Это связано с недавними терактами в этом городе.

Одна из американских фирм по [\textcolor{Emerald}{их}]$_{SA}$ производству сообщила, что в офисе не прекращают звонить телефоны. Владельцы квартир в высотных зданиях интересуются возможностью приобретения [\textcolor{Turquoise}{новой модели парашюта}]$_1$, [\textcolor{Turquoise}{которая}]$_1$ была разработана после трагических событий 11 сентября. [\textcolor{Turquoise}{Он}]$_1$ стоит около 800 долларов и раскрывается автоматически. [...]
} &

Residents of New York skyscrapers rush to get 

[\textcolor{Emerald}{parachutes}]$_{SA}$

Residents of skyscrapers in New York rush to get

[\textcolor{Emerald}{parachutes}]$_{SA}$. This is due to recent terrorist attacks in the city.

One of [\textcolor{Emerald}{their}]$_{SA}$ manufacturer reports that the phones in its office never stop ringing. Apartment owners in high-rise buildings are interested in buying [\textcolor{Turquoise}{a new parachute}]$_1$ [\textcolor{Turquoise}{which}]$_1$ is developed after the tragic events of September 11. [\textcolor{Turquoise}{It}]$_1$ costs about \$800 and opens automatically. [...]
\\
\bottomrule
\end{tabular}
\caption{\label{tab:sa_example} Split-antecedent annotation example in the RuCoCo dataset, from \texttt{2001\_world\_new\_003}. }
\vspace{-3\baselineskip}%
\end{table}

We perform the experiments on the RuCoCo-2023 Shared Task dataset described in \cite{dobrovolskii2022rucoco}. It is a large corpus for coreference resolution collected from news articles in Russian. It contains annotated news in multiple categories, including finance, world news, sports, and more. The corpus includes both single one-to-one coreference annotation and split antecedents one-to-many coreference annotation. However, the distinguishing feature of the latter is that it is annotated among clusters (entities), not mentions (an example is shown in Table~\ref{tab:sa_example}). It poses a challenge in identifying pairs of mentions from different groups that are connected by split-antecedent relations. To address this additional challenge, our model's architecture would require additional modifications. Although both tasks are evaluated jointly in the competition, this study's emphasis is on the standard coreference resolution. Here, we conduct some additional analyses of the data relevant to our methods.

\begin{figure}[ht]
	\centering
	\begin{minipage}{0.45\textwidth}
	\centering
	\includegraphics[width=0.8\textwidth]{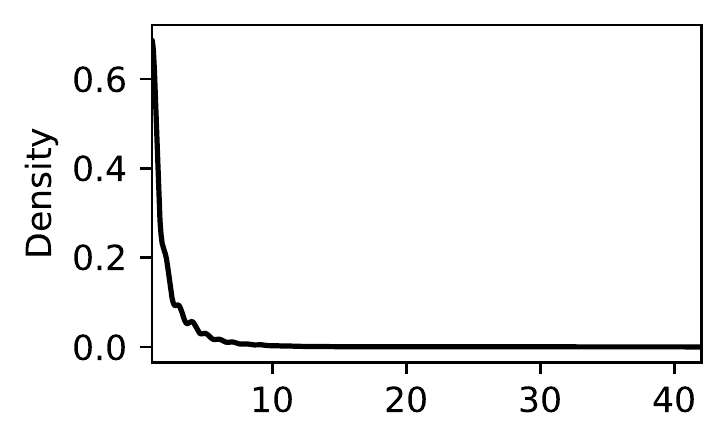}
    \caption{Entity lengths in tokens (train set).}
    \label{fig:entity_lengths}
	\end{minipage}
	\hfill
	\begin{minipage}{0.47\textwidth}
	\centering
    \includegraphics[width=0.8\textwidth]{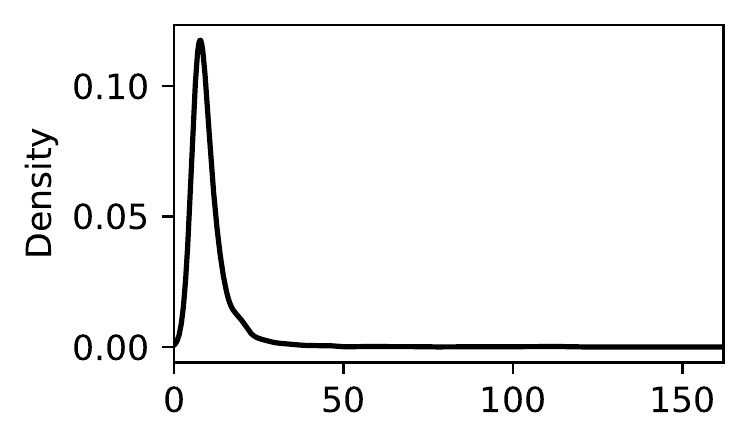}
    \caption{Number of paragraphs per annotation.}
    \label{fig:paragraph_lengths}
	\end{minipage}
\end{figure}

Firstly, it is critical for our model to determine the maximum entity length in the corpus. The results on the train set are illustrated in Fig. \ref{fig:entity_lengths}. The mean entity length is 2, and the maximum is 42. The maximum mention length in our system is set to 13 tokens, which covers 99.7\% of entities in the corpus.

Secondly, we examine the number of paragraphs in the data. It will be identical to the number of trees in RST parser output. Thus, if we construct the text-level tree by merging paragraph trees, it could be critical for long discourse dependencies. The results are shown in Fig. \ref{fig:paragraph_lengths}. 
Every line split is considered a paragraph. The median paragraph count is 9, with the maximum number of separated lines being~162. Some news articles are exceptionally long, and some of them include enumerated lists. Combining multiple trees into one can affect the referential distance estimation in a few particularly long texts.

\subsection{Evaluation}
\label{ssec:evaluation}

In the Shared Task, the coreference resolution F1 score is calculated using the Link-based Entity Aware (LEA) metric \cite{moosavi-strube-2016-coreference}. In this metric, the weight of each entity is determined by its size, with larger entities being considered more important. It also evaluates resolved coreference relations instead of resolved mentions.

The models are validated during training using 5\% of the official train set. We run random splitting 4 times and report the average result. The listed results on the official development and test sets of the competition are obtained with the exact same models.

\section{Results and Discussion}
\label{sec:results}

In Table \ref{tab:main-results}, we present the results of our experiments on the development set of the RuCoCo-2023 Shared Task. We also report the performance of our system on the test set (also called the final set) of the RuCoCo-2023 Shared Task in Table \ref{tab:test-results}. Our baseline model noticeably outperforms the \texttt{RuRoBERTa-large}-based baseline provided by the organizers, which achieved 68.4\% and 67.4\% F1 on the development and test sets, respectively.

\begin{table}[ht]
 \centering
\begin{tabular}{llllc}
\toprule
                    & \bf Precision  & \bf Recall   & \bf F1       & \bf \begin{tabular}[c]{@{}l@{}}Top-1 F1\\ (leaderboard)\end{tabular}\\ \midrule
Baseline            & 78.7 ± 0.7     & 69.1 ± 0.7   & 73.5 ± 0.5  & 74.3 \\
$+D_{Lin}$          & 78.6 ± 1.8     & 68.3 ± 2.2   & 73.0 ± 0.5  & 74.0 \\
$+D_{Rh}$           & 78.5 ± 1.5     & 69.3 ± 1.0   & \bf 73.6 ± 0.9  & \bf 74.6 \\
$+D_{LCA}$          & 75.0 ± 0.8     & 70.9 ± 1.0   & 72.9 ± 0.4  & 73.5 \\

\bottomrule
\end{tabular}
\caption{Models evaluation on the official development set.}
\label{tab:main-results}
\end{table}

Due to the strict limit on the number of submissions in the final phase of the competition, we could only evaluate the two best performing models, Baseline and Baseline$+D_{Rh}$, on a private leaderboard.

\begin{table}[ht]
 \centering
\begin{tabular}{rlllc}
\toprule
                    & \bf Precision  & \bf Recall   & \bf F1        & \bf \begin{tabular}[c]{@{}l@{}}Top-1 F1\\ (leaderboard)\end{tabular}\\ \midrule
Baseline            & 79.1 ± 0.8    & 66.9 ± 0.6    & \bf 72.5 ± 0.3 & 72.8     \\
$+D_{Rh}$           & 79.3 ± 1.6    & 66.6 ± 1.9    & 72.4 ± 0.5     & \bf 73.3 \\

\bottomrule
\end{tabular}
\caption{Models evaluation on the official test set (``Final'').}
\label{tab:test-results}
\end{table}

Features $D_{Lin}$ and $D_{LCA}$ are not found to be effective for the task of neural coreference resolution on the development set (Table \ref{tab:main-results}). Our hypothesis is that $D_{Lin}$, the linear distance in elementary DUs, may not offer much more information than the linear distance in tokens that the neural model already employs. $D_{LCA}$, the distance from the right-hand mention to the LCA, on the other hand, may not be accurate when we artificially merge the RST trees for each paragraph into a single right-branched tree. In this case, the depth of the right-hand branch depends more on the order of paragraphs than the actual discourse structure of the text.

The mean results of the model enhanced with the rhetorical distances $D_{Rh}$ are not much different from the baseline results on both sets. However, its results vary more, hence the model with the best F1 score reached both leaderboards. This suggests to us that the rhetorical distance is more robust than the other features, even though it shares all the mentioned drawbacks of the other features.

\section{Conclusion}
\label{sec:conclusion}

In this paper, we propose a new method for neural coreference resolution that incorporates discourse information. We test our method on the RuCoCo-2023 Shared Task and demonstrate that it outperforms the competition baseline by a significant margin, while also ranking 1st on the development set and 2nd on the test set of the competition. The key findings of this work are:

\begin{enumerate}
    \item We implemented various features related to distances in the text-level RST tree to study how the hierarchical discourse information obtained with discourse parser can help coreference resolution for Russian.
    \item We observed a marginal improvement using the rhetorical distance feature. The model that uses this feature got the best result on the Shared Task development and test sets.
    \item We used the multilingual entity-aware LUKE model and showed that it performs competitively with the monolingual language models for Russian in coreference resolution, even with limited computational resources.
\end{enumerate}

These findings suggest that the multilingual entity-aware LUKE model is a viable option for coreference resolution in Russian, and despite the constraints of the current rhetorical analyzer for Russian that prevent full-text analysis, the features of hierarchical discourse can still be found useful. We hope that our work will inspire further research on incorporating referential distance information into neural coreference resolution models.

\section*{Acknowledgements}

The research was carried out using the infrastructure of the Shared Research Facilities ``High Performance Computing and Big Data'' (CKP ``Informatics'') of FRC CSC RAS (Moscow). This study was conducted within the framework of the scientific program of the National Center for Physics and Mathematics, section №9 ``Artificial intelligence and big data in technical, industrial, natural and social systems''.

\bibliography{dialogue,anthology}
\bibliographystyle{dialogue}

\end{document}